\documentclass[10pt,twocolumn,letterpaper]{article}

\usepackage[pagenumbers]{cvpr} 

\definecolor{cvprblue}{rgb}{0.21,0.49,0.74}
\usepackage[pagebackref,breaklinks,colorlinks,allcolors=cvprblue]{hyperref}
\usepackage{comment}
\usepackage{algorithm}
\usepackage{algpseudocode}
\usepackage{graphicx}
\usepackage{array} 

\def\paperID{1903} 
\def\confName{CVPR}
\def\confYear{2026}

\title{2K-Characters-10K-Stories: A Quality-Gated Stylized Narrative Dataset with Disentangled Control and Sequence Consistency}

\author{Xingxi Yin \quad Yicheng Li \quad Gong Yan \quad Chenglin Li \\
 Zhejiang University\\
 {\tt\small \{yinxingxi,liyicheng,gongyan,lichenglin\}@zju.edu.cn}
\and
Jian Zhao \quad Cong Huang \quad Yue Deng \\
Zhongguancun Institute of Artificial Intelligence\\
 {\tt\small \{zhaojian, huangcong,dengyue\}@zgci.ac.cn}
\and
Yin Zhang\\
 Zhejiang University\\
 {\tt\small zhanyin@zju.edu.cn}
}

\begin{document}

\twocolumn [{%
\renewcommand\twocolumn [1] [] {#1}%
\maketitle
\begin{center}
\centering
\includegraphics[width=1\textwidth]{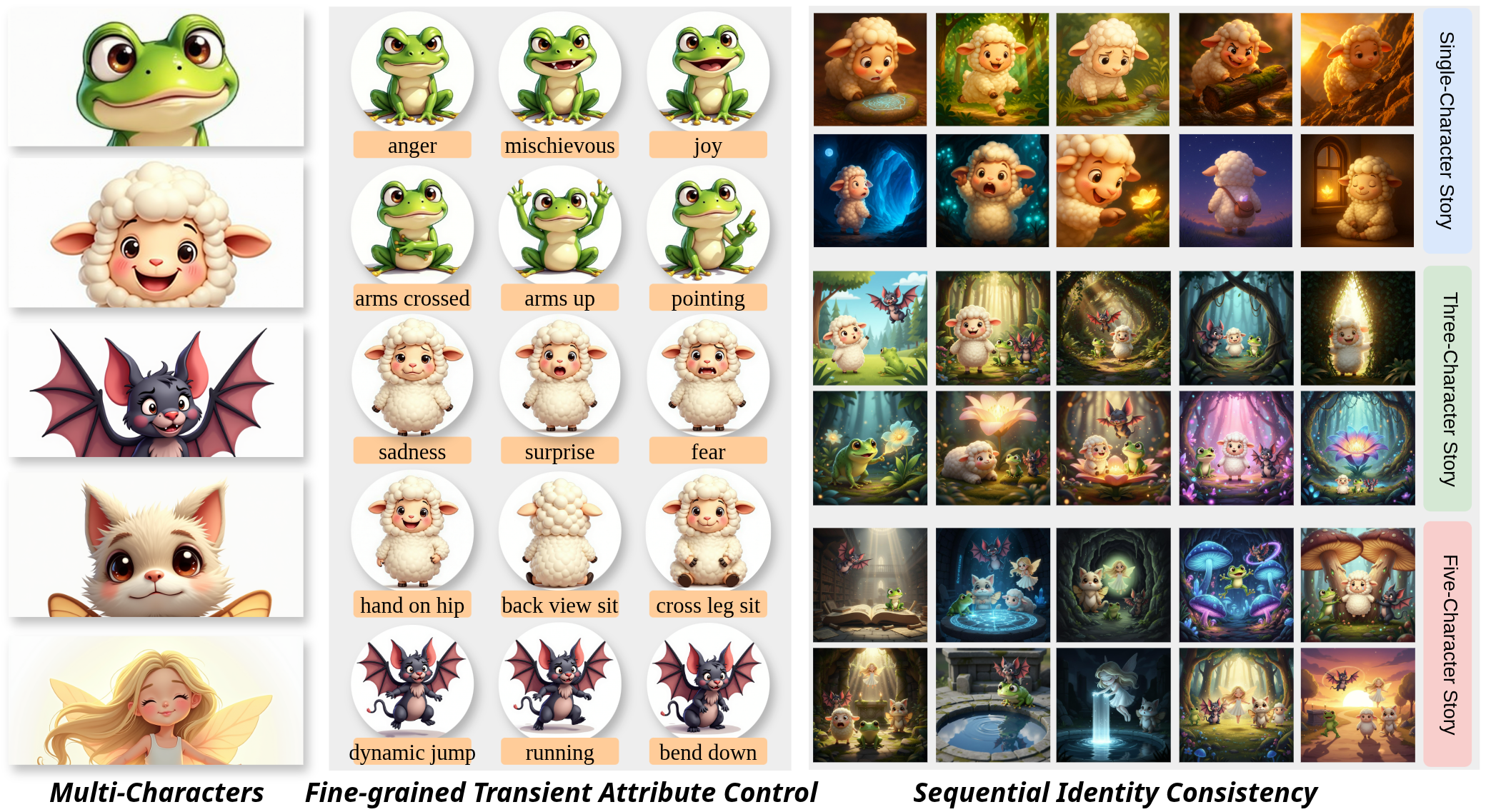}


\captionof{figure}{\textbf{Our Structured HiL Pipeline synthesizes high-fidelity visual narratives.} The left panel illustrates the multi-character identity templates. The middle panel enables precise manipulation of transient attributes for a character identity. The right panels demonstrate robust sequential identity preservation across complex visual narratives, validating the fidelity of our Quality-Gated synthesis process.}
\label{fig:teaser}
\end{center}%
}]

\begin{abstract}

Sequential identity consistency under precise transient attribute control remains a long-standing challenge in controllable visual storytelling. Existing datasets lack sufficient fidelity and fail to disentangle stable identities from transient attributes, limiting structured control over pose, expression, and scene composition and thus constraining reliable sequential synthesis. To address this gap, we introduce \textbf{2K-Characters-10K-Stories}, a multi-modal stylized narrative dataset of \textbf{2{,}000} uniquely stylized characters appearing across \textbf{10{,}000} illustration stories. It is the first dataset that pairs large-scale unique identities with explicit, decoupled control signals for sequential identity consistency. We introduce a \textbf{Human-in-the-Loop pipeline (HiL)} that leverages expert-verified character templates and LLM-guided narrative planning to generate highly-aligned structured data. A \textbf{decoupled control} scheme separates persistent identity from transient attributes---pose and expression---while a \textbf{Quality-Gated loop} integrating MMLM evaluation, Auto-Prompt Tuning, and Local Image Editing enforces pixel-level consistency. Extensive experiments demonstrate that models fine-tuned on our dataset achieves performance comparable to closed-source models in generating visual narratives.

\end{abstract}    
\section{Introduction}
\label{sec:intro}

The creation of controllable visual content and the advancement of Visual Storytelling remain central challenges in generative AI. Despite the impressive fidelity achieved by modern Text-to-Image (T2I) models \cite{rombach2022high,saharia2022photorealistic,openai2023dalle3,Midjourney} and personalized generation techniques \cite{gal2022image,ruiz2023dreambooth,kumari2023multi,wang2024ms,kang2025flux}, extending single-image generation capabilities to multi-frame, multi-character narratives introduces fundamentally new complexities: \textbf{identity consistency across sequential narrative story frames} \cite{hu2024animate,mao2024story,ren2023make,zhuang2025vistorybench,zhou2024storydiffusion,oliveira2025storyreasoning,yang2024seedstory,cheng2024theatergen,mao2024story_adapter,Liu_2024_CVPR} and \textbf{fine-grained transient attribute control} \cite{zhang2023adding,mou2024t2i} during the generation process. The scalability and generalization of identity preservation represents the first bottleneck. Dedicated identity techniques, while successful for individual subjects, fail to maintain consistency across large character sets and generalize due to the inherent limitations in available training data. Concurrently, the second bottleneck is the ambiguity of structural control. Specialized fine-grained attribute manipulation methods (e.g. ControlNet \cite{zhang2023adding}, SPACE \cite{gururani2023space}), while effective for isolated controls, fundamentally lack the identity anchoring and narrative context necessary for control transfer across a story sequence. This deficiency hinders the shift from vague prompt engineering to precise, quantifiable manipulation.

Critically, existing datasets fundamentally lack the structural organization and scale required to address these two bottlenecks. Identity-focused story visualization datasets (e.g., VIST \cite{huang2016visual}, StorySalon \cite{liu2024intelligent}, Flintstones \cite{gupta2018imagine}, PororoSV \cite{li2019storygan}, OpenStory \cite{ye2024openstory} and StoryStream \cite{yang2024seedstory}) are highly constrained, lacking in identity scale and diversity and fundamentally missing decoupled annotations for dynamic attributes such as pose and expression, as summarized in Table \ref{tab:t1}. This inability to direct the generative process with discrete control signals limits their utility for research requiring fine-grained, quantifiable synthesis. This pervasive \textbf{data structure gap}—missing both scale and \textbf{the combination of fine-grained descriptions and rigorously aligned target images}—is the fundamental barrier preventing current SOTA models from achieving coherent, controllable visual narratives. 

We address this critical gap by introducing \textbf{2K-Characters-10K-Stories}, a large-scale, multi-modal \textbf{stylized narrative dataset} specifically designed for controllable visual illustration. This dataset is built to enable the study of both \textbf{stylistic narrative consistency} and \textbf{fine-grained transient attribute control}. We constructed this dataset using a rigorous, advanced \textbf{Human-in-the-Loop (HiL)} pipeline that features an \textbf{iterative quality-gating loop}. This loop is driven by MMLM Auto-Evaluation and leverages both \textbf{Auto-Prompt Tuning (APT)} and precision-guided \textbf{Local Image Editing (LIE)} to enforce a decoupled binding between character identity and transient attributes, thereby ensuring both structural rigor and ultra-high data fidelity across the entire sequence. 

Our core contributions are summarized as follows: 
\begin{enumerate}
  \item \textbf{Advanced structured human-in-the-Loop pipeline.} We design a subject-driven HiL pipeline that combines expert-verified character templates, LLM-generated scripts, and an iterative quality-gating loop with multi-stage human validation to produce highly consistent narrative illustrations.
  \item \textbf{Granular decoupled control.} We provide explicit, decoupled control signals for every character instance, transforming character generation from ad-hoc prompt tuning into precise, quantifiable control over identity, pose, expression, and composition.
  \item \textbf{Sequential consistency under fine-grained control.} We build the first public dataset that couples sequential identity consistency with fine-grained transient attribute control, covering 2,000 expert-verified stylized characters across 10,000+ multi-panel stories (75,000+ illustrations). Models trained on this dataset achieve substantial gains in character consistency and control fidelity.
\end{enumerate}

\section{Related works}
\label{sec:related_work}

\begin{table}
    \centering
\begin{tabular}{ccc>{\centering\arraybackslash}p{0.15\linewidth}}
 Dataset&Characters&Stories& DCS\\
    \midrule
    VIST&Dispersed&$\sim$40K& $\times$\\
 Flintstones&$\le 7$&$\sim$25K& $\times$\\
 ProroSV&$\le 13$&$\sim$15K& $\times$\\
 StorySalon&446&$\sim$11K& $\times$\\
 StoryStream&$\le 15$&Not specified& $\times$\\
 OpenStory& Dispersed& Not specified& $\times$\\
    Ours&2000&10K& $\checkmark$\\
      \bottomrule
\end{tabular}
  \caption{Fundamental shortcomings of state-of-the-art datasets for Controllable Multi-Character Narrative Synthesis, focusing on their limited Unique Character/Story Scale and the lack of explicit, independent control signals (e.g., Pose/Expression). DCS represent Disentangled Control Signals.}
  \label{tab:t1}
\end{table}

Our work contributes to the fields of \textbf{visual narrative synthesis}, \textbf{scalable identity preservation}, and \textbf{structurally controllable generative modeling}. To contextualize the necessity of our \textbf{2K-Characters-10K-Stories} dataset and address the pervasive data structure gap, we critically review existing literature across three key areas.

\paragraph{Missing Decoupled Supervision in Visual Narrative Datasets.} Early efforts in visual storytelling, encompassing datasets such as VIST \cite{huang2016visual}, Flintstones \cite{gupta2018imagine}, and PororoSV \cite{li2019storygan}, primarily focused on establishing basic semantic alignment and narrative coherence (e.g., text-to-image matching). However, these resources are fundamentally limited by their confined scale and structural ambiguity, failing to provide the explicit annotations required for quantifiable control. This structural deficiency severely restricts the development of robust, generalizable identity features and, crucially, prevents fine-grained manipulation of transient attributes such as precise expression, posture, and action consistency across story sequences. Identity-focused benchmarks, such as ViStoryBench \cite{zhuang2025vistorybench}, highlight the consistency consistency problem but are constrained by their limited subject count (e.g., 344 unique characters). The central failure across all these existing works is the absence of the requisite scale for training decoupled models, alongside a lack of explicit, semantic-level decoupled supervision. Our work uniquely bridges this gap by offering 2,000+ unique subjects and a large-scale set of rigorously aligned fine-grained transient attributes descriptions, effectively transitioning the control task from subjective description to precise, objective synthesis control. 

\paragraph{Controllable Generation and Mechanism-Level Limitations.} The challenge of reliable character control exposes the mechanism-level limitations of current generative models regarding ID preservation and structural attribute manipulation. \textbf{For identity preservation}, state-of-the-art personalized T2I methods (e.g., DreamBooth \cite{ruiz2023dreambooth},  IP-Adapter \cite{ye2023ip} ) struggle in continuous narratives. Without explicit supervision, their ID embeddings subtly couple with the text's dynamic instructions, preventing the decoupling of ID features from transient attributes (e.g., pose/expression). This coupling leads to cumulative drift in generalization across frames.\textbf{For structural control}, precise manipulation of pose and expression is critical. While conditioning mechanisms like ControlNet \cite{zhang2023adding} offer geometric guidance, existing T2I pipelines rely on continuous embedding spaces from large text encoders (e.g., CLIP \cite{radford2021learning} and T5 \cite{raffel2020exploring}) for high-level control instructions. This reliance on continuous natural language features leads to semantic ambiguity and compromises precision, as descriptive text features tend to entangle and interfere with each other, resulting in a discrepancy between the intended description and the final generated image \cite{chang2022semanticaware}.

\paragraph{Data-Centric AI and Automated Quality-Gating.} The fidelity of large-scale generative datasets is highly dependent on rigorous curation methodologies. Modern Data-Centric AI principles \cite{wang2023data} necessitate advanced quality control that goes beyond simple filtering. Our work advances this methodology by presenting a novel iterative quality-gating pipeline that ensures both scale and precision. This framework integrates MMLMs (e.g., GPT-4o \cite{hurst2024gpt} or Gemini \cite{gemini2023family} via API) for automated, objective MMLM Auto-Evaluation and script refinement \cite{liu2023visual}. We further introduce Auto-Prompt Tuning  \cite{ramnath2025systematic,hao2023optimizing,yang2024ampo,mo2024dynamic,li2025survey} for optimizing generation success rates and precision-guided Local Image Editing (LIE) \cite{wu2025qwenimagetechnicalreport} to correct localized inconsistencies without compromising global coherence. This comprehensive, AI-assisted approach ensures that our final 75,000+ images meet an unprecedented standard of visual fidelity and structural accuracy, providing a critically needed benchmark for future research in mechanism-level generative control.

\section{Method}
\label{sec:method}
\begin{figure*}[!ht]
    \centering
    \includegraphics[width=1.0\textwidth]{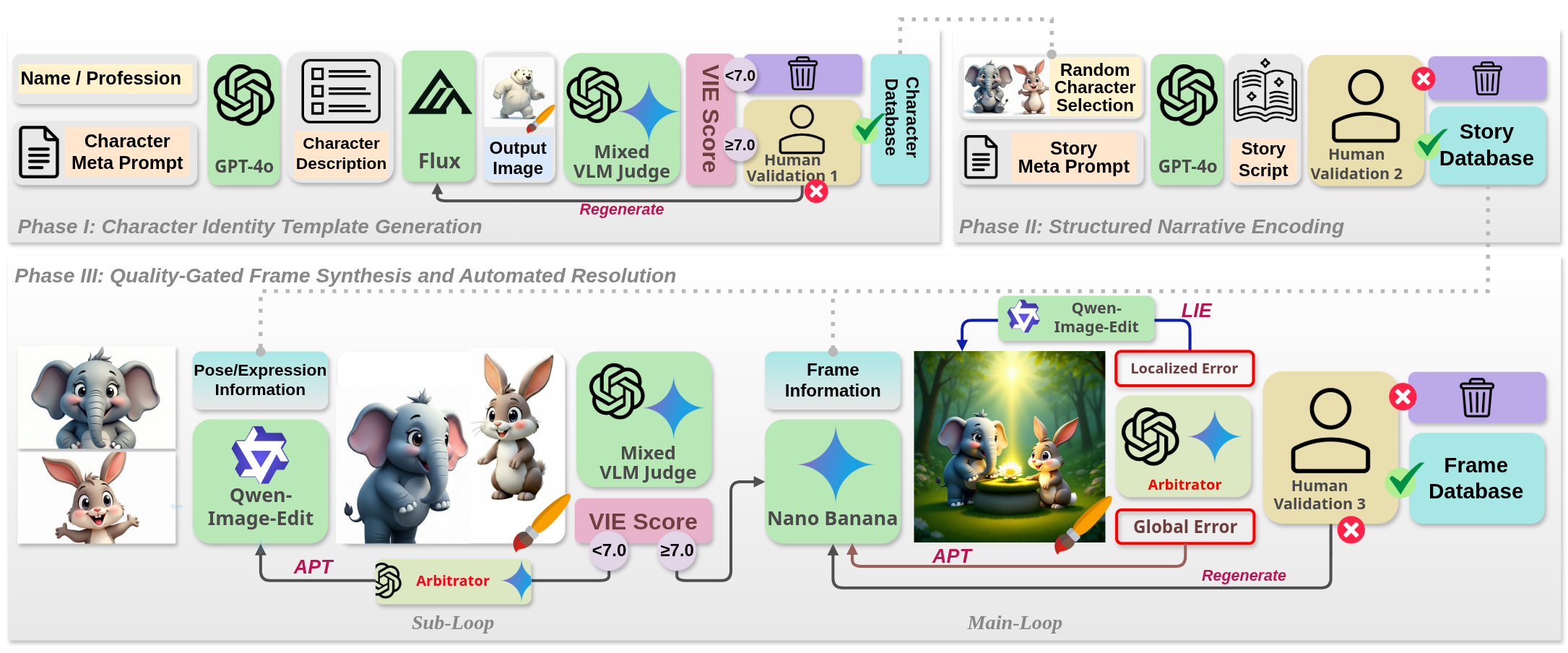}

    \caption{\textbf{Three-Phase Human-in-the-Loop (HiL) Pipeline for Quality-Gated, Controllable Narrative Data Synthesis.} This HiL pipeline enforces semantic control and quality assurance across four phases. \textbf{Phase I (Design)} establishes visual constraints, including Character ID templates (Human Validation 1). \textbf{Phase II (Encoding)} transforms narrative text into the Structured Frame Metadata (Human Validation 2). The core generation process occurs in \textbf{Phase III (Synthesis and Resolution)}. After image generation, the output is immediately routed through the \textbf{Integrated Quality Control and Correction (IQCC)}. The IQCC unifies the Triple-Check diagnosis and initiates a targeted automated resolution strategy (APT or LIE) based on the scope and type of error.} 
    \label{fig:pipeline}
\end{figure*}

\begin{figure}
    \centering
    \includegraphics[width=\linewidth]{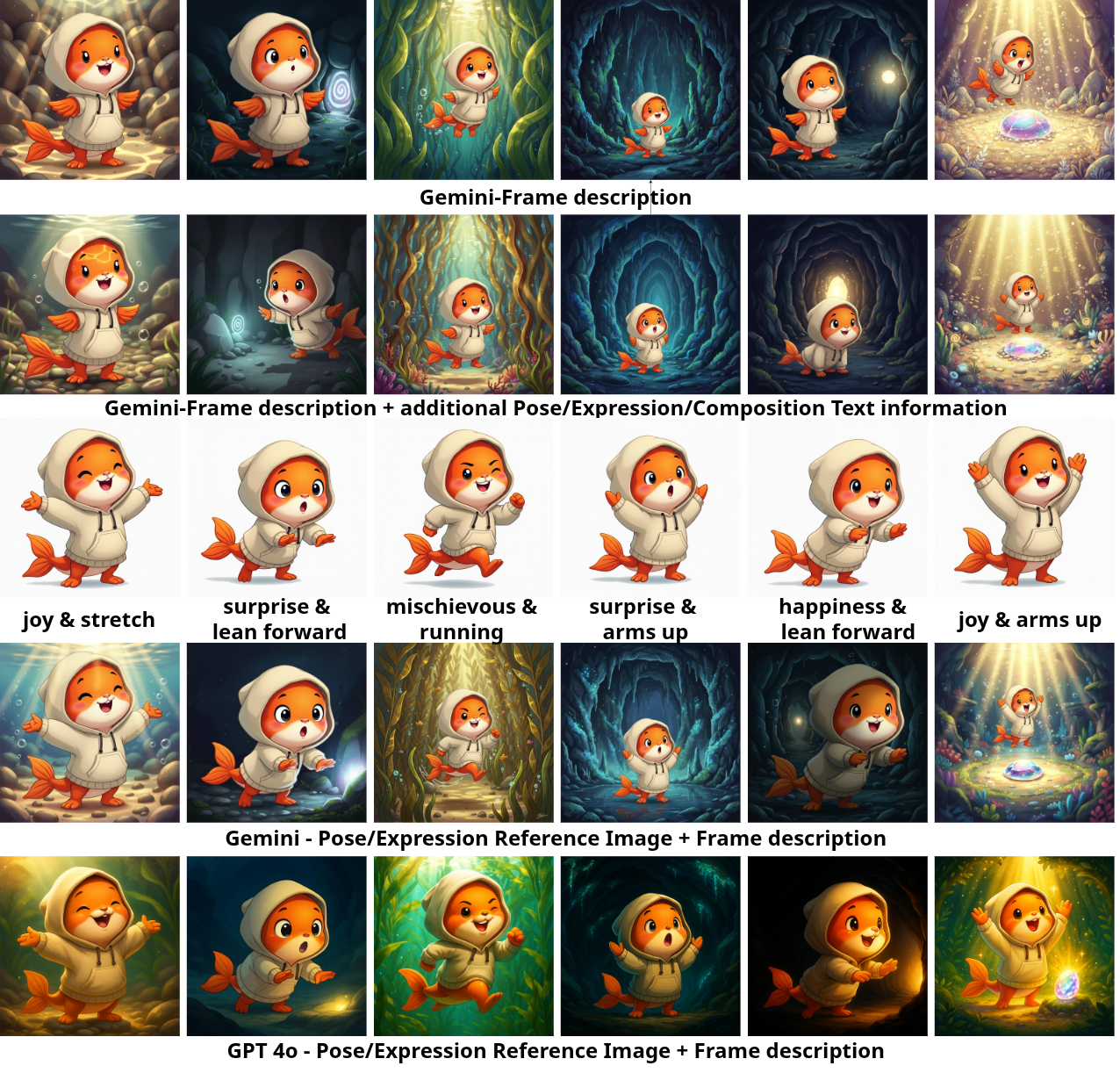}
    \caption{\textbf{Ablation Study of data Synthesis Strategies.} This comparison demonstrates the necessity of our HiL control mechanism before the IQCC process. \textbf{Row 1 (Text-Only)} shows that raw frame descriptions result in low fidelity and uncontrolled attribute variation. \textbf{Row 2 (Text Augmented)} shows marginal improvements, but still lacks the geometric precision required for complex control. Critically, \textbf{Rows 4 and 5} illustrate the adopted two-step strategy: using a pose/expression reference image as a dedicated control signal is essential for achieving stable, high-fidelity $C_{ID}$ preservation and accurate transient attribute control.} 
    \label{fig:data_generate_ablation}
\end{figure}

\begin{figure}
    \centering
    \includegraphics[width=\linewidth]{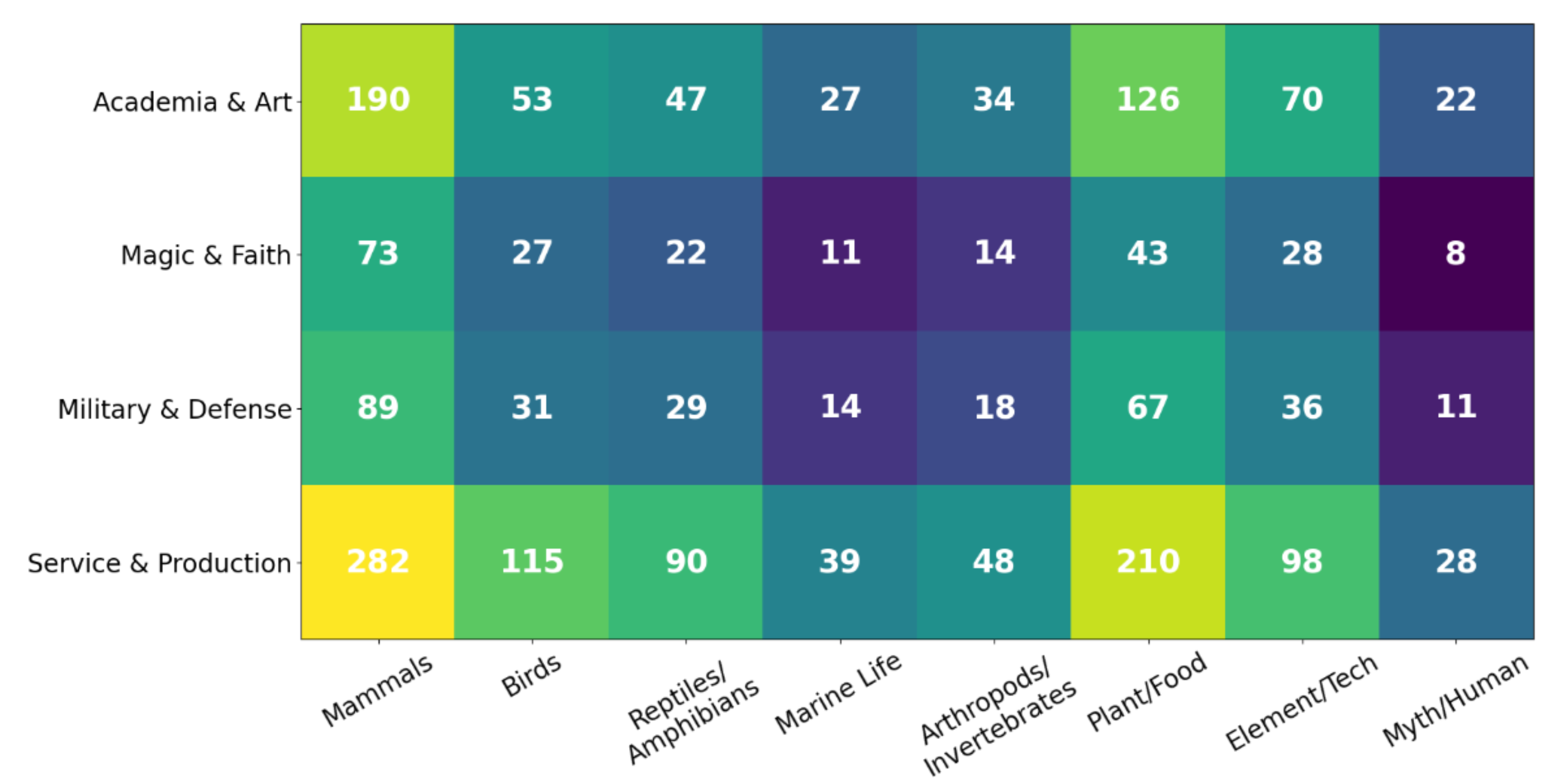}
    \caption{\textbf{Joint Distribution of 2K Characters by Subject-Type and Profession.} This concise heatmap visualizes the character pool's diversity across eight distinct subject-types (X-axis) and four distinct profession categories (Y-axis). The values represent the raw count for each intersection, confirming a robust and balanced representation of characters.}
    \label{fig:characters_statistic}
\end{figure}

\begin{figure}
    \centering
    \includegraphics[width=\linewidth]{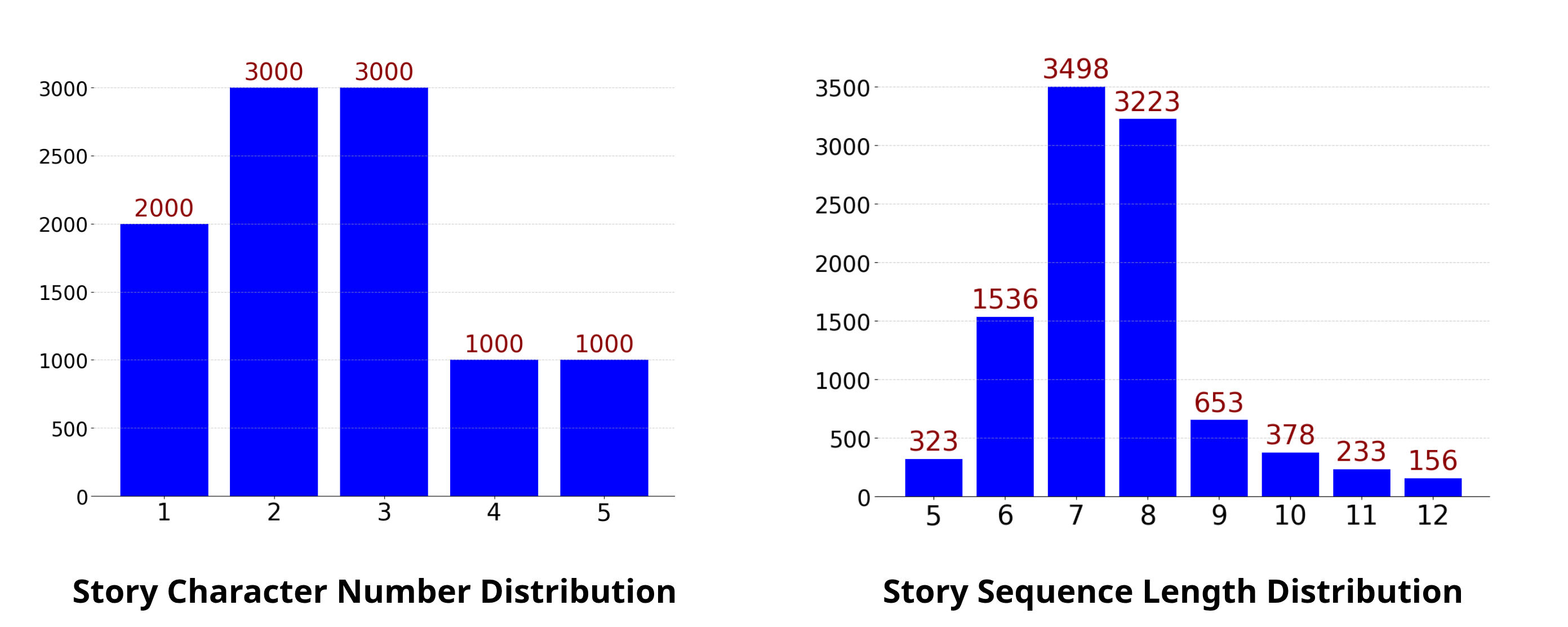}
    \caption{\textbf{Statistical Distribution of 10K-Stories Complexity.} This figure validates the dataset's scale and structured design. (Left) The Character Number Distribution is intentionally balanced across 1 to 5 characters, ensuring broad social complexity. (Right) The Sequence Length Distribution confirms the dataset's focus on sequential consistency, primarily featuring stories between 6 and 9 frames.} 
    \label{fig:story_distribution}
\end{figure}

The construction of the stylized narrative dataset is built via a rigorous three-phase Structured Human-in-the-Loop pipeline (illustrated in Fig.~\ref{fig:pipeline}).  This pipeline sequentially handles Character Identity (Phase I), Structured Narrative Encoding (Phase II), and final Quality-Gated Frame Synthesis and Automated Resolution (Phase III), ensuring explicit decoupling and verifiable attribute fidelity in every frame. This methodology is engineered to enforce \textbf{sequential identity consistency} and achieve \textbf{quantifiable, fine-grained control} over character transient attributes. To mitigate inherent model biases, the core LLM/MLLM stages implement a Model Ensembling Strategy, randomly invoking the Gemini Pro 2.5 \cite{gemini2023family} and GPT-4o \cite{openai2023gpt4} APIs with an equal chance. The efficacy of our pipeline is fundamentally built upon a novel scheme for decoupling character identity from transient attributes, which we detail first. 

\textbf{Structured Data and Decoupled Control.} The core innovation lies in the use of Semantic Identifiers for Decoupled Control. All control signals ($P_{ID}$, $E_{ID}$, $C_{TAG}$) are unique, semantically meaningful text identifiers, guaranteeing high  extensibility and semantic transparency (see Fig.~\ref{fig:data_generate_ablation}). The control space is rigorously defined by \textbf{28 distinct Pose IDs}, \textbf{12 distinct Expression IDs}, and \textbf{21 distinct Composition Tags}. The definition of the decoupled control space is given in Appendix A. 

\subsection{Phase I: Character Identity Template Generation} This phase establishes the immutable identity anchors for each character. We begin by generating character descriptions based on the Character Meta Prompt (detailed in Appendix A) and the character's MLLM-derived subject type and profession, which are then visualized with Flux \cite{BlackForestLabs2024Flux1Dev}. The resulting character description and image are initially evaluated via VIEScore \cite{ku2023viescore}.  Any concepts scoring below 7.0 are immediately discarded. Concepts that pass are then subject to Human Validation 1 (HV1) by a three-expert panel, requiring a consensus of two or more affirmative votes to lock the Char ID. Every successful $C_{ID}$ image is standardized to  $1024  * 1024$ pixels as a full-body representation against a plain white background for maximum clarity and minimal feature interference.  This rigorous process yielded a final role-playing character dataset of 2,000 unique character images. Specifically, the dataset features unprecedented fine-grained diversity, including over \textbf{650 unique subject types} (e.g., mammals, mythology, plants, marine life, and humans) and over \textbf{440 unique professions} (covering knowledge/academia/art, magic/mystery/belief, military/defense, and daily services). The joint statistical distribution is given in Fig.~\ref{fig:characters_statistic}, and the detailed information and additional character samples are provided in Appendix B.

\subsection{Phase II: Structured Narrative Encoding.} This stage focuses on script planning and structuring, converting high-level, character-centric concepts into precise, structured metadata, leveraging multimodal input for automated narrative generation. The LLM ensemble is provided with the following inputs:  a) The visual template files ($C_{ID}$ images) for a set of 1 to 5 randomly sampled characters; b) A generic Story Meta Prompt (textual input) defining the genre. The visual descriptions of the sampled characters from Phase I are automatically extracted and used to overwrite the corresponding generic character placeholders within this text prompt, which explicitly includes the list of semantic identifiers as strict generation constraints. The model is strictly constrained to generate a sequence of 7 to 12 sequential frames.  The details of the Story Meta Prompt are given in Appendix C. The LLM's core function \cite{gemini2023family,hurst2024gpt} is to act as a Structured Encoder. Its output is a Structured JSON Object containing the complete narrative text and the required control signals. For each frame $i$ in the story, the output structure $S_ i$ is explicitly defined:
\[
S_i = \{ \text{Frame Text}_i, \{C_{ID_j}, P_{ID_j}, E_{ID_j}, C_{TAG_j}\}_i \} 
\]
This ensures that each character $j$ has its identity referenced via  the $C_{ID_j}$, while transient attributes are systematically varied according to the plot.  A sample of the generated structured JSON object (Story Script) is provided in Appendix C.  Expert reviewers then conduct Human Validation 2 (HV2) to audit the scripts, verifying the narrative coherence and the semantic justification for the $P_{ID}$, $E_{ID}$ and $C_{TAG}$ transitions of each character across the sequence. Approval for the script requires a two-vote consensus from the panel.

\subsection{Phase III: Quality-Gated Frame Synthesis and Automated Resolution.}  This phase deploys a \textbf{Nested Iterative Loop} for rigorous, pixel-level enforcement of consistency. We adopt a two-stage generation strategy since the ablation study (illustrated in Fig.~\ref{fig:data_generate_ablation}) confirms that direct prompting with frame descriptions (row 1) or augmented text (row 2) leads to poor pose and expression fidelity. Our two-step approach (rows 4 and 5) first translates semantic constraints into a pixel-level reference image, which provides the necessary hard structural control for the final frame synthesis.

\paragraph{Sub-Loop: High-Fidelity Reference Generation.} We initiate a sub-loop to generate high-fidelity references $R_{ij}$ for each character $j$ present in the frame, fusing identity $C_{ID_j}$ with transient attributes (Pose and Expression) using a dedicated image editing model \cite{wu2025qwenimagetechnicalreport}. This is critical as it explicitly decouples $ P_{ID}$ and $ E_{ID}$ into a verifiable visual anchor.  The quality is quantitatively measured by the VIEScore \cite{ku2023viescore}. References are only approved if the  VIEScore exceeds 7.0. Failure triggers \textbf{Auto-Prompt Tuning (APT)}: the MMLM generates a textual diagnostic report and iteratively optimizes the prompt (e.g., via keyword weighting or restructuring) before regeneration. The efficiency of the APT mechanism is high: for those reference images that failed the VIEScore quality gate, APT requires an average of only \textbf{2.19 iterations} to achieve a successful modification. The detailed APT Meta Prompt and statistical information are provided in Appendix D.

\paragraph{Main Loop: Reference-Guided Frame Synthesis and Triple-Check Evaluation. } The validated reference image ($R_{ij}$) and the frame description prompt drive the Main Loop synthesis via Gemini-2.5-flash-image (Nano Banana) \cite{gemini2023family} , which is capable of image generation. The MLLM uses these $R_{ij}$ as the primary visual control signal to integrate all characters into the final scene, strictly adhering to the background and composition constraints specified in the text prompt. The resulting image is immediately subjected to the Triple-Check Automated Evaluation by a separate MMLM Arbitrator \cite{gemini2023family,openai2023gpt4} , which verifies three criteria: (i) Identity Consistency, (ii) Control Accuracy (Pose/Expression), and (iii) Semantic Alignment (Context/Composition). The subsequent evaluation and \textbf{Automated Failure Correction} are unified under the \textbf{Integrated Quality Control and Correction (IQCC)}, executed by the MMLM arbitrator \cite{gemini2023family,openai2023gpt4}. The detailed Meta Prompt governing IQCC's JSON output and decision logic is provided in Appendix D. The IQCC performs a triple-check validation against all semantic and visual constraints. A detected failure immediately initiates a targeted resolution strategy, mapped directly based on the severity and scope of the error:
\begin{enumerate}
    \item \textbf{Global Error.} If the failure is due to identity consistency, semantic alignment, or composition/viewpoint issues, the system triggers Auto-Prompt Tuning. The MMLM generates an optimized prompt for complete frame regeneration. 
    \item \textbf{Localized Error.}  If the failure is due to Control Accuracy or minor object/color inconsistencies, the system triggers Local Image Editing. The MMLM provides precise and concise instructions for targeted, localized repairs using high-performance editing models  (e.g., Qwen \cite{wu2025qwenimagetechnicalreport}).
    \item \textbf{Hard Failure.} If the APT or LIE sequence fails to achieve automated approval after 3 consecutive attempts. The sample is flagged as a Hard Failure and is automatically escalated to \textbf{Human Validation 3 (HV3)} for manual inspection and final disposition. 
\end{enumerate}

HV3 serves as the ultimate arbitration layer, exclusively handling hard failures \textbf{that persist after 3 consecutive APT/LIE attempts}. \textbf{Approximately 2\% of generated samples escalate to this process (detailed in Appendix D).} The final disposition—manual correction, full regeneration, or discard—is determined by the expert panel's mandatory two-vote consensus, ensuring maximum data integrity. The  distribution of story complexity is given in Fig.~\ref{fig:story_distribution}.  Three story samples are provided in Fig.~\ref{fig:teaser}, while more story examples are available in Appendix D, which includes styles such as \textbf{fantasy concept}, \textbf{Pixar}, \textbf{illustration style}, and \textbf{long-range story samples}, \textbf{etc.} 
\section{Experiment}
\label{sec:experiment}
\begin{figure*}[!ht]
    \centering
    \includegraphics[width=1.0\textwidth]{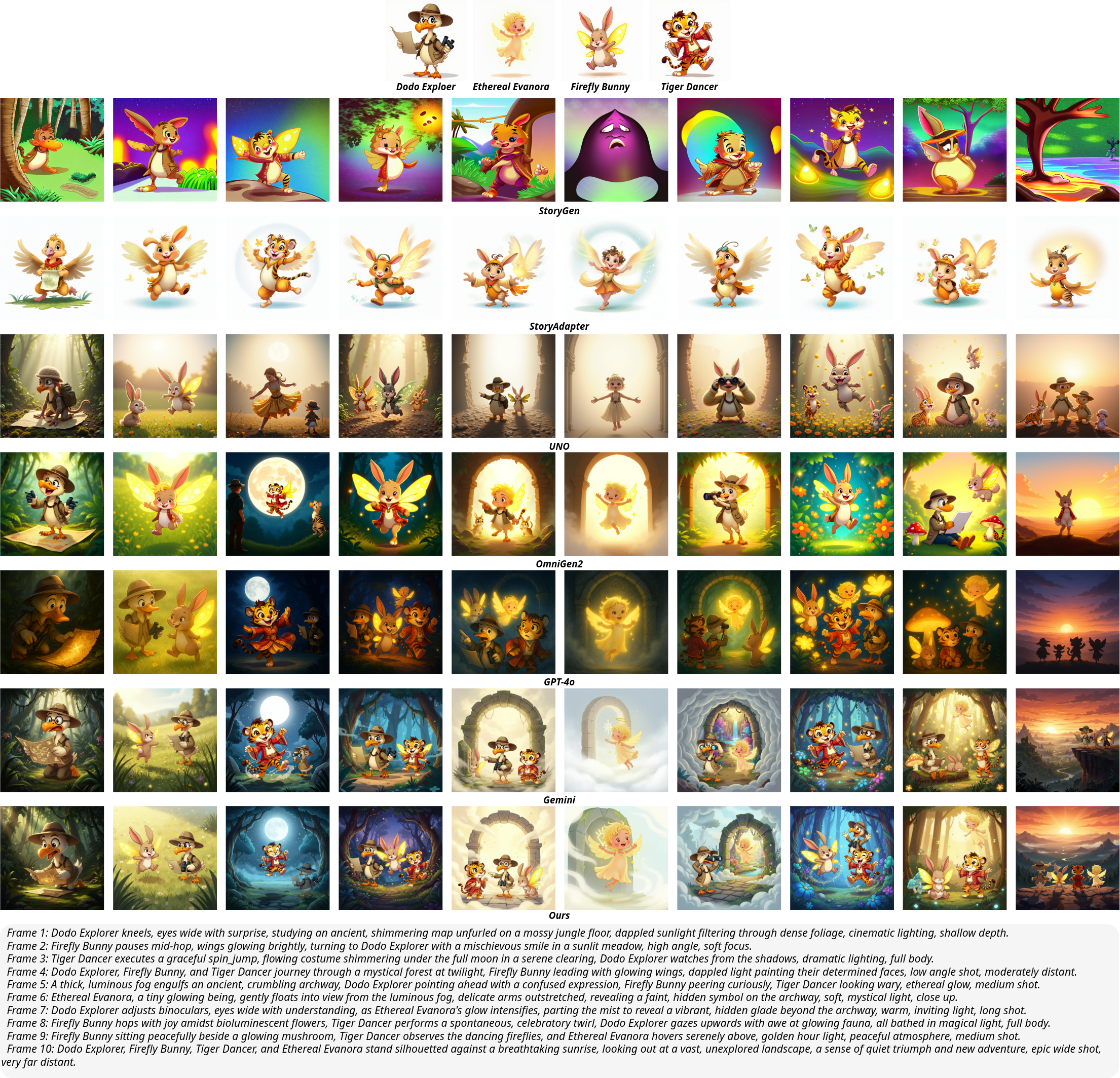}

    \caption{\textbf{Qualitative Validation of the 2K-Characters-10K-Stories Dataset.} This figure presents the qualitative generation results for 10-frame extended story sequences driven by four distinct and contrasting characters. This qualitative analysis highlights two core advantages resulting from training on our dataset: (1) Robust cross-frame identity coherence maintained across multi-character and long sequences; and (2) Accurately executed fine-grained geometric and semantic controls (pose, expression, and composition) compared to all baselines.} 
    \label{fig:quality_experiment}
\end{figure*}

\begin{table*}

    \centering
\begin{tabular}{c>{\centering\arraybackslash}p{0.06\linewidth}>{\centering\arraybackslash}p{0.06\linewidth}>{\centering\arraybackslash}p{0.06\linewidth}>{\centering\arraybackslash}p{0.06\linewidth}>{\centering\arraybackslash}p{0.05\linewidth}>{\centering\arraybackslash}p{0.04\linewidth}>{\centering\arraybackslash}p{0.04\linewidth}>{\centering\arraybackslash}p{0.05\linewidth}c>{\centering\arraybackslash}p{0.07\linewidth}}
    \toprule
    & \multicolumn{2}{c}{\underline{\hspace{0.1cm}CSD (Style)\hspace{0.1cm}}}&\multicolumn{2}{c}{\underline{\hspace{0.1cm}CIDS (Character)\hspace{0.1cm}}}&  \multicolumn{4}{c}{\underline{\hspace{1.2cm}Alignment Score\hspace{1.2cm}}}& Inception&Aesthetics\\
 Method& $Cross\uparrow$& $Self\uparrow$& $Cross\uparrow$& $Self\uparrow$&   $Scene\uparrow$& $pose\uparrow$& $Exp\uparrow$&$Comp\uparrow$& score$\uparrow$&score$\uparrow$\\
    \midrule
    storyGen&  52.928 & 66.499&  63.314& 72.642
&   29.43& 21.75& 28.28&25.33& 4.15&4.649
\\
 storyAdapter& \textbf{65.777}& \textbf{84.62}& 62.619& 72.082&   25.73& 17.60& 20.45&20.39
& 3.65&5.406
\\
 UNO& 56.579& 72.871& 66.128& 72.442&   80.33& 48.10& 44.08&41.15
& \textbf{6.67}&5.881
\\
 OmniGen2& 58.893& 66.219& 68.039& 73.831
&   73.90& 38.03& 45.25&36.68& 4.01&5.851
\\
 GPT4o& 57.039& 68.471& 66.242& 73.765
&   93.10& 65.15& 65.50&62.00& 4.20&
5.835
\\
 Gemini& 59.788& 69.841& \textbf{69.741}& 74.315
&   \textbf{95.25}& 61.45& 70.38&58.78& 4.56&\textbf{6.122}\\
    Ours& 60.449& 68.308&   69.611& \textbf{75.844}&   90.45& \textbf{70.70}& \textbf{77.10}&\textbf{66.25}& 4.33&6.054
\\
      \bottomrule
\end{tabular}
  \caption{\textbf{Comprehensive Quantitative Evaluation Across Structural Control and Sequential Coherence.}  The table presents a comprehensive quantitative comparison across all baseline models and our fine-tune model on all defined metrics. \textbf{Narrative Coherence} is measured by $CSD$, $CIDS$ and Scene Alignment. \textbf{Control Fidelity} is measured by Pose/Expression/Composition Alignment score. \textbf{Image Quality} is measured by Inception Score and Aesthetics Score.All metrics are reported as higher-is-better ($\uparrow$)}
  \label{tab:t2}
\end{table*}

Our experiments are constructed to provide comprehensive quantitative and qualitative evidence for the efficacy of high-fidelity, structurally aligned training data in complex sequential narrative generation tasks. While story visualization can involve various subjects, this work focuses on the challenging domain of controllable stylized narrative with model-generated virtual characters. This specific task exposes two fundamental bottlenecks in multi-frame, multi-character narratives: 1) maintaining identity consistency across sequential narrative story frames, and 2) achieving fine-grained structural narrative control during the generation process. We aim to validate that training on the quality-gated and structurally aligned \textbf{2K-Characters-10K-Stories Dataset}  is an essential prerequisite for effectively resolving these two challenges within this specialized domain.

\paragraph{Model Training and Validation.} We utilize the custom \textbf{2K-Characters-10K-Stories Dataset}, derived from our Human-in-the-Loop pipeline, as the foundation for our experiments. This dataset is partitioned into a training set comprising over 75,000 IQCC-verified frames, where each sample is a highly-aligned triplet ($C_{ID}$ Template images, Story frame text prompt, Target Frame Image). We adopt the OmniGen2 architecture \cite{wu2025omnigen2} for its native multi-control channel capabilities, allowing it to seamlessly integrate both the character identity image and the structured frame prompt essential for complex sequential control. The model is fine-tuned on the training set using OmniGen2's default training parameters and scheduler configuration for reproducibility. Crucially, the auxiliary control signal ($R_{ij}$), which was used for data quality assurance in Phase III IQCC, is explicitly omitted from the training input. This ensures that the model achieves geometric control solely through the structured frame text prompt. For rigorous validation, the evaluation set consists of \textbf{500 ten-frame stories (5,000 contiguous frames)}, featuring \textbf{200 Out-of-Distribution characters} to stringently test our model's generalization capability.

\paragraph{Baseline models and Evaluation Metrics.} For a fair and relevant comparison, we benchmark our approach against six strong state-of-the-art baselines: StoryGen \cite{Liu_2024_CVPR} , StoryAdapter \cite{mao2024story_adapter}, UNO \cite{wu2025less}, OmniGen2 \cite{wu2025omnigen2}, Gemini \cite{gemini2023family}, and GPT-4o \cite{openai2023gpt4}. We deliberately exclude direct comparison with models such as LatentUnfold \cite{kang2025latentunfold}, StoryDiffusion \cite{zhou2024storydiffusion}, SEED-Story \cite{yang2024seedstory}, and TheaterGen \cite{cheng2024theatergen} due to fundamental task misalignment and input constraints. Our task necessitates generating story frames conditioned on multiple character reference images and a corresponding scene description. However, models like LatentUnfold \cite{kang2025latentunfold}, StoryDiffusion \cite{zhou2024storydiffusion}, and  SEED-Story \cite{yang2024seedstory} cannot support our multi-character input requirement, as they are limited to one or two input images. Furthermore, TheaterGen \cite{cheng2024theatergen} demands extra Bounding Box annotations and complex decomposition of the frame prompt, which is incompatible with our structured input.  Regarding evaluation, we adopt established metrics from ViStoryBench \cite{zhuang2025vistorybench}, including CSD-based style similarity (CSD), Character Identification Similarity (CIDS), Inception Score, and Aesthetics Score. Critically, we enhance the standard Alignment Score by decomposing it into four distinct, fine-grained metrics: the scene alignment score and the character-specific pose, expression, and composition alignment scores. These decomposed metrics provide a detailed, separate assessment of the text-image consistency and character attribute accuracy, demonstrating the efficacy of our structural alignment in complex controlled generation. 

\paragraph{Results and Analysis.} The quantitative evaluation results are presented comprehensively in Table \ref{tab:t2}, comparing our fine-tuned model, the base OmniGen2 \cite{wu2025omnigen2}, and all baseline methods across Control Fidelity, Narrative Coherence, and General Image Quality Metrics.  \textbf{Validation of Data Alignment Quality.} The primary finding confirms the necessity and profound impact of the high-fidelity, structurally aligned data provided by our HiL pipeline. Our fine-tuned model, trained on the IQCC-verified data, substantially outperforms the ablation model (base OmniGen2 \cite{wu2025omnigen2}) in all Control Fidelity metrics. This is evidenced by significant improvements in pose alignment ($+32.65$), expression alignment ($+31.85$) composition alignment ($+29.57$), and frame scene alignment ($+16.55$) (refer to Table \ref{tab:t1}). This conclusive performance gap validates that the rigorous alignment between image content and frame text is the \textbf{key differentiator} for achieving fine-grained, unambiguous control over character attributes. \textbf{Superior Coherence and Domain Quality.} The quantitative results demonstrate comprehensive performance gains across all measured metrics compared to the base OmniGen2, including style similarity (CSD), character identity identification (CIDS), Inception Score ($+0.312$), and Aesthetics Score ($+0.203$). Furthermore, in the specialized domain of stylized narrative, our model's output rivals the Narrative Coherence of powerful closed-source models like Gemini \cite{gemini2023family} and GPT 4o \cite{openai2023gpt4}. Crucially, our model significantly surpasses these closed-source models in all Control Fidelity metrics (e.g., Pose/Expression/Composition alignment scores). Further validation on the public ViStoryBench benchmark (Table 3, Supp. Mat.) confirms the structural efficacy of our data. All results indicate that our structurally curated data effectively endows an open-source architecture with state-of-the-art domain-specific quality, surpassing both general T2I baselines and models dedicated solely to control.

\paragraph{Case Study.}We present a detailed qualitative comparison in Fig.~\ref{fig:quality_experiment} , illustrating the generation results across a challenging 10-frame narrative sequence featuring four contrasting characters: Dodo Explorer, Ethereal Evanora, Firefly Bunny, and Tiger Dancer. This visual evidence strongly corroborates our quantitative findings and highlights the unique advantages imparted by our highly aligned dataset. \textbf{Identity Coherence and Sequence Stability.} With changes in character interactions and environmental settings over the sequence (e.g., Frame 2-5), most baseline models (StoryGen, storyAdapter, and UNO) suffer from severe identity drift and style instability, fundamentally failing to maintain the visual fidelity of characters throughout the narrative. While the closed-source models GPT-4o, Gemini exhibit high single-frame quality, they display noticeable inconsistencies in character details. For instance, in Frames 3 and 8, the visual representation of Tiger Dancer appears to be a direct copy of the reference image, thus failing to seamlessly integrate the character into the generated scene's lighting and style. Furthermore, they often fail to correctly position all four characters in challenging compositional frames (e.g., Frame 10). In sharp contrast, our fine-tuned model successfully preserves the nuanced identity, color scheme, and texture of all four characters consistently across the entire 10-frame sequence. \textbf{Fine-grained Control Execution.} The reference text prompts require complex, frame-level control over pose, expression, and multi-character composition (e.g., Frame 4: `Tiger Dancer turns journey through a mystical forest at twilight'). Baselines often struggle to realize these precise instructions; for instance, OmniGen2 frequently suffers from fundamental errors such as omitting characters, merging identities, or failing to render specific poses. A clear example is Frame 4, where the model fails to include both Dodo Explorer and Tiger Dancer. Our fine-tuned model, however, demonstrates the most accurate execution of the scene's semantic and geometric controls, precisely capturing the specified actions, emotions, and complex group compositions, thus proving the effectiveness of training on our structurally aligned data for precise conditional generation.

\section{Conclusion}
\label{sec:conclusion}
In this work, we presented a novel  Human-in-the-Loop (HiL) data construction pipeline to generate the \textbf{2K-Characters-10K-Stories Dataset}. This structured approach conclusively validates that data quality and explicit alignment of transient attributes are essential prerequisites for precise conditional generation in sequential narratives, successfully addressing bottlenecks in identity coherence and fine-grained control over character transient attributes. Our experiments showed that training on this IQCC-verified data significantly boosts the performance of an open-source architecture (OmniGen2), yielding substantial improvements in Control Fidelity metrics (+32.65) in pose alignment and enabling it to surpass powerful closed-source baselines in transient attribute control. While the HiL pipeline is inherently resource-intensive, limiting its immediate scalability, this framework establishes a critical quality standard for controllable generation. Future work will concentrate on automating the quality assurance steps to enhance pipeline scalability and expanding the dataset to incorporate temporal emotion and motion controls, ultimately extending its application to high-fidelity video generation.


\newpage

{
    \small
    \bibliographystyle{ieeenat_fullname}
    \bibliography{main}

@String(CVPR= {IEEE Conf. Comput. Vis. Pattern Recog.})

@String(ECCV= {Eur. Conf. Comput. Vis.})

@String(AAAI = {AAAI})

@String(CVPR  = {CVPR})

@String(ECCV  = {ECCV})

@article{raffel2020exploring,
  title={Exploring the limits of transfer learning with a unified text-to-text transformer},
  author={Raffel, Colin and Shazeer, Noam and Roberts, Adam and Lee, Katherine and Narang, Sharan and Matena, Michael and Zhou, Yanqi and Li, Wei and Liu, Peter J},
  journal={Journal of machine learning research},
  volume={21},
  number={140},
  pages={1--67},
  year={2020}
}

@inproceedings{zhang2023adding,
  title={Adding conditional control to text-to-image diffusion models},
  author={Zhang, Lvmin and Rao, Anyi and Agrawala, Maneesh},
  booktitle={Proceedings of the IEEE/CVF International Conference on Computer Vision},
  pages={3836--3847},
  year={2023}
}

@article{wu2025omnigen2,
  title={OmniGen2: Exploration to Advanced Multimodal Generation},
  author={Wu, Chenyuan and Zheng, Pengfei and Yan, Ruiran and Xiao, Shitao and Luo, Xin and Wang, Yueze and Li, Wanli and Jiang, Xiyan and Liu, Yexin and Zhou, Junjie and others},
  journal={arXiv preprint arXiv:2506.18871},
  year={2025}
}

@article{ye2023ip,
  title={{IP-Adapter}: Text Compatible Image Prompt Adapter for Text-to-Image Diffusion Models},
  author={Ye, Hu and Zhang, Jun and Liu, Sibo and Han, Xiao and Yang, Wei},
  journal={arXiv preprint arXiv:2308.06721},
  year={2023}
}

@inproceedings{chang2022semanticaware,
  title={{Semantic-aware Contrastive Learning for More Accurate Semantic Parsing}},
  author={Chang, Shuaichen and Peng, Jiajie and Wang, Binghao and Li, Yizhi and Zhou, Jingkun and Wang, Jingjing and Wang, Yihong},
  booktitle={Proceedings of the 2022 Conference on Empirical Methods in Natural Language Processing (EMNLP)},
  pages={2693--2704},
  year={2022}
}

@article{liu2023visual,
  title={{Visual Instruction Tuning}},
  author={Liu, Haotian and Li, Chunyuan and Wu, Qingyang and Lee, Yong Jae},
  journal={arXiv preprint arXiv:2304.08485},
  year={2023}
}

@article{gemini2023family,
  title={{Gemini: A Family of Highly Capable Multimodal Models}},
  author={{Gemini Team} and {Google}},
  journal={arXiv preprint arXiv:2312.11805},
  year={2023},
  url={https://arxiv.org/abs/2312.11805}
}

@article{ramnath2025systematic,
  title={A systematic survey of automatic prompt optimization techniques},
  author={Ramnath, Kiran and Zhou, Kang and Guan, Sheng and Mishra, Soumya Smruti and Qi, Xuan and Shen, Zhengyuan and Wang, Shuai and Woo, Sangmin and Jeoung, Sullam and Wang, Yawei and others},
  journal={arXiv preprint arXiv:2502.16923},
  year={2025}
}

@article{hao2023optimizing,
  title={Optimizing prompts for text-to-image generation},
  author={Hao, Yaru and Chi, Zewen and Dong, Li and Wei, Furu},
  journal={Advances in Neural Information Processing Systems},
  volume={36},
  pages={66923--66939},
  year={2023}
}

@InProceedings{Liu_2024_CVPR,
        author    = {Liu, Chang and Wu, Haoning and Zhong, Yujie and Zhang, Xiaoyun and Wang, Yanfeng and Xie, Weidi},
        title     = {Intelligent Grimm - Open-ended Visual Storytelling via Latent Diffusion Models},
        booktitle = {Proceedings of the IEEE/CVF Conference on Computer Vision and Pattern Recognition (CVPR)},
        month     = {June},
        year      = {2024},
        pages     = {6190-6200}
}

@misc{mao2024story_adapter,
  title={{Story-Adapter: A Training-free Iterative Framework for Long Story Visualization}},
  author={Mao, Jiawei and Huang, Xiaoke and Xie, Yunfei and Chang, Yuanqi and Hui, Mude and Xu, Bingjie and Zhou, Yuyin},
  journal={arXiv},
  volume={abs/2410.06244},
  year={2024},
}

@article{yang2024ampo,
  title={Ampo: Automatic multi-branched prompt optimization},
  author={Yang, Sheng and Wu, Yurong and Gao, Yan and Zhou, Zineng and Zhu, Bin Benjamin and Sun, Xiaodi and Lou, Jian-Guang and Ding, Zhiming and Hu, Anbang and Fang, Yuan and others},
  journal={arXiv preprint arXiv:2410.08696},
  year={2024}
}

@article{cheng2024theatergen,
  title={TheaterGen: Character Management with LLM for Consistent Multi-turn Image Generation},
  author={Cheng, Junhao and Yin, Baiqiao and Cai, Kaixin and Huang, Minbin and Li, Hanhui and He, Yuxin and Lu, Xi and Li, Yue and Li, Yifei and Cheng, Yuhao and others},
  journal={arXiv preprint arXiv:2404.18919},
  year={2024}
}

@inproceedings{li2019storygan,
  title={Storygan: A sequential conditional gan for story visualization},
  author={Li, Yitong and Gan, Zhe and Shen, Yelong and Liu, Jingjing and Cheng, Yu and Wu, Yuexin and Carin, Lawrence and Carlson, David and Gao, Jianfeng},
  booktitle={Proceedings of the IEEE/CVF conference on computer vision and pattern recognition},
  pages={6329--6338},
  year={2019}
}

@inproceedings{gupta2018imagine,
  title={Imagine this! scripts to compositions to videos},
  author={Gupta, Tanmay and Schwenk, Dustin and Farhadi, Ali and Hoiem, Derek and Kembhavi, Aniruddha},
  booktitle={Proceedings of the European conference on computer vision (ECCV)},
  pages={598--613},
  year={2018}
}

@inproceedings{liu2024intelligent,
  title={Intelligent Grimm -- Open-ended Visual Storytelling via Latent Diffusion Models}, 
  author={Chang Liu, Haoning Wu},
  booktitle={The IEEE/CVF Conference on Computer Vision and Pattern Recognition (CVPR)},
  year={2024},
}

@article{kang2025latentunfold,
      title={Flux Already Knows - Activating Subject-Driven Image Generation without Training}, 
      author={Kang, Hao and Fotiadis, Stathi and Jiang, Liming and Yan, Qing and Jia, Yumin and Liu, Zichuan and Chong, Min Jin and Lu, Xin},
      journal={arXiv preprint}, 
      volume={arXiv:2504.11478},
      year={2025},
}

@article{wu2025less,
  title={Less-to-More Generalization: Unlocking More Controllability by In-Context Generation},
  author={Wu, Shaojin and Huang, Mengqi and Wu, Wenxu and Cheng, Yufeng and Ding, Fei and He, Qian},
  journal={arXiv preprint arXiv:2504.02160},
  year={2025}
}

@inproceedings{ye2024openstory,
  title={Openstory: A large-scale open-domain dataset for subject-driven visual storytelling},
  author={Ye, Zilyu and Liu, Jinxiu and Cao, JinJin and Chen, Zhiyang and Xuan, Ziwei and Zhou, Mingyuan and Liu, Qi and Qi, Guo-Jun},
  booktitle={Proceedings of the IEEE/CVF Conference on Computer Vision and Pattern Recognition},
  pages={7953--7962},
  year={2024}
}

@article{yang2024seedstory,
      title={SEED-Story: Multimodal Long Story Generation with Large Language Model}, 
      author={Shuai Yang and Yuying Ge and Yang Li and Yukang Chen and Yixiao Ge and Ying Shan and Yingcong Chen},
      year={2024},
      journal={arXiv preprint arXiv:2407.08683},
      url={https://arxiv.org/abs/2407.08683}, 
}

@article{li2025survey,
  title={A survey of automatic prompt engineering: An optimization perspective},
  author={Li, Wenwu and Wang, Xiangfeng and Li, Wenhao and Jin, Bo},
  journal={arXiv preprint arXiv:2502.11560},
  year={2025},
}

@inproceedings{mo2024dynamic,
  title={Dynamic prompt optimizing for text-to-image generation},
  author={Mo, Wenyi and Zhang, Tianyu and Bai, Yalong and Su, Bing and Wen, Ji-Rong and Yang, Qing},
  booktitle={Proceedings of the IEEE/CVF Conference on Computer Vision and Pattern Recognition},
  pages={26627--26636},
  year={2024}
}

@article{wang2023data,
  title={{Data Management For Training Large Language Models: A Survey}},
  author={Wang, Zige and Zhong, Wanjun and Wang, Yufei and Zhu, Qi and Mi, Fei and Wang, Baojun and Shang, Lifeng and Jiang, Xin and Liu, Qun},
  journal={arXiv preprint arXiv:2312.01700},
  year={2023}
}

@article{saharia2022photorealistic,
  title={Photorealistic text-to-image diffusion models with deep language understanding},
  author={Saharia, Chitwan and Chan, William and Saxena, Saurabh and Li, Lala and Whang, Jay and Denton, Emily L and Ghasemipour, Kamyar and Gontijo Lopes, Raphael and Karagol Ayan, Burcu and Salimans, Tim and others},
  journal={Advances in neural information processing systems},
  volume={35},
  pages={36479--36494},
  year={2022}
}

@inproceedings{ruiz2023dreambooth,
  title={Dreambooth: Fine tuning text-to-image diffusion models for subject-driven generation},
  author={Ruiz, Nataniel and Li, Yuanzhen and Jampani, Varun and Pritch, Yael and Rubinstein, Michael and Aberman, Kfir},
  booktitle={Proceedings of the IEEE/CVF Conference on Computer Vision and Pattern Recognition},
  pages={22500--22510},
  year={2023}
}

@misc{wu2025qwenimagetechnicalreport,
      title={Qwen-Image Technical Report}, 
      author={Chenfei Wu and Jiahao Li and Jingren Zhou and Junyang Lin and Kaiyuan Gao and Kun Yan and Sheng-ming Yin and Shuai Bai and Xiao Xu and Yilei Chen and Yuxiang Chen and Zecheng Tang and Zekai Zhang and Zhengyi Wang and An Yang and Bowen Yu and Chen Cheng and Dayiheng Liu and Deqing Li and Hang Zhang and Hao Meng and Hu Wei and Jingyuan Ni and Kai Chen and Kuan Cao and Liang Peng and Lin Qu and Minggang Wu and Peng Wang and Shuting Yu and Tingkun Wen and Wensen Feng and Xiaoxiao Xu and Yi Wang and Yichang Zhang and Yongqiang Zhu and Yujia Wu and Yuxuan Cai and Zenan Liu},
      year={2025},
      eprint={2508.02324},
      archivePrefix={arXiv},
      primaryClass={cs.CV},
      url={https://arxiv.org/abs/2508.02324}, 
}

@article{ku2023viescore,
  title={Viescore: Towards explainable metrics for conditional image synthesis evaluation},
  author={Ku, Max and Jiang, Dongfu and Wei, Cong and Yue, Xiang and Chen, Wenhu},
  journal={arXiv preprint arXiv:2312.14867},
  year={2023}
}

@misc{BlackForestLabs2024Flux1Dev,
  author = {{Black Forest Labs}},
  title = {{FLUX.1-dev}},
  howpublished = {\url{https://huggingface.co/black-forest-labs/FLUX.1-dev}},
  year = {2024},
  note = {Open-weight, guidance-distilled model for non-commercial applications; Accessed: 2025-07-06}
}

@inproceedings{mou2024t2i,
  title={T2i-adapter: Learning adapters to dig out more controllable ability for text-to-image diffusion models},
  author={Mou, Chong and Wang, Xintao and Xie, Liangbin and Wu, Yanze and Zhang, Jian and Qi, Zhongang and Shan, Ying},
  booktitle={Proceedings of the AAAI Conference on Artificial Intelligence},
  volume={38},
  number={5},
  pages={4296--4304},
  year={2024}
}

@inproceedings{radford2021learning,
  title={Learning transferable visual models from natural language supervision},
  author={Radford, Alec and Kim, Jong Wook and Hallacy, Chris and Ramesh, Aditya and Goh, Gabriel and Agarwal, Sandhini and Sastry, Girish and Askell, Amanda and Mishkin, Pamela and Clark, Jack and others},
  booktitle={International conference on machine learning},
  pages={8748--8763},
  year={2021},
  organization={PMLR}
}

@inproceedings{rombach2022high,
  title={High-resolution image synthesis with latent diffusion models},
  author={Rombach, Robin and Blattmann, Andreas and Lorenz, Dominik and Esser, Patrick and Ommer, Bj{\"o}rn},
  booktitle={Proceedings of the IEEE/CVF conference on computer vision and pattern recognition},
  pages={10684--10695},
  year={2022}
}

@inproceedings{hu2024animate,
  title={Animate anyone: Consistent and controllable image-to-video synthesis for character animation},
  author={Hu, Li},
  booktitle={Proceedings of the IEEE/CVF Conference on Computer Vision and Pattern Recognition},
  pages={8153--8163},
  year={2024}
}

@article{mao2024story,
  title={Story-adapter: A training-free iterative framework for long story visualization},
  author={Mao, Jiawei and Huang, Xiaoke and Xie, Yunfei and Chang, Yuanqi and Hui, Mude and Xu, Bingjie and Zhou, Yuyin},
  journal={arXiv preprint arXiv:2410.06244},
  year={2024}
}

@article{gal2022image,
  title={An image is worth one word: Personalizing text-to-image generation using textual inversion},
  author={Gal, Rinon and Alaluf, Yuval and Atzmon, Yuval and Patashnik, Or and Bermano, Amit H and Chechik, Gal and Cohen-Or, Daniel},
  journal={arXiv preprint arXiv:2208.01618},
  year={2022}
}

@inproceedings{huang2016visual,
title={Visual storytelling},
author={Huang, Ting-Hao and Ferraro, Francis and Mostafazadeh, Nasrin and Misra, Ishan and Agrawal, Aishwarya and Devlin, Jacob and Girshick, Ross and He, Xiaodong and Kohli, Pushmeet and Batra, Dhruv and others},
booktitle={Proceedings of the 2016 conference of the North American chapter of the association for computational linguistics: Human language technologies},
pages={1233--1239},
year={2016}
}

@article{oliveira2025storyreasoning,
  title={StoryReasoning Dataset: Using Chain-of-Thought for Scene Understanding and Grounded Story Generation},
  author={Oliveira, Daniel AP and de Matos, David Martins},
  journal={arXiv preprint arXiv:2505.10292},
  year={2025}
}

@inproceedings{gururani2023space,
  title={Space: Speech-driven portrait animation with controllable expression},
  author={Gururani, Siddharth and Mallya, Arun and Wang, Ting-Chun and Valle, Rafael and Liu, Ming-Yu},
  booktitle={Proceedings of the ieee/cvf international conference on computer vision},
  pages={20914--20923},
  year={2023}
}

@article{zhou2024storydiffusion,
  title={Storydiffusion: Consistent self-attention for long-range image and video generation},
  author={Zhou, Yupeng and Zhou, Daquan and Cheng, Ming-Ming and Feng, Jiashi and Hou, Qibin},
  journal={Advances in Neural Information Processing Systems},
  volume={37},
  pages={110315--110340},
  year={2024}
}

@article{zhuang2025vistorybench,
  title={Vistorybench: Comprehensive benchmark suite for story visualization},
  author={Zhuang, Cailin and Huang, Ailin and Cheng, Wei and Wu, Jingwei and Hu, Yaoqi and Liao, Jiaqi and Wang, Hongyuan and Liao, Xinyao and Cai, Weiwei and Xu, Hengyuan and others},
  journal={arXiv preprint arXiv:2505.24862},
  year={2025}
}

@article{ren2023make,
  title={Make-a-character: High quality text-to-3d character generation within minutes},
  author={Ren, Jianqiang and He, Chao and Liu, Lin and Chen, Jiahao and Wang, Yutong and Song, Yafei and Li, Jianfang and Xue, Tangli and Hu, Siqi and Chen, Tao and others},
  journal={arXiv preprint arXiv:2312.15430},
  year={2023}
}

@article{hurst2024gpt,
  title={Gpt-4o system card},
  author={Hurst, Aaron and Lerer, Adam and Goucher, Adam P and Perelman, Adam and Ramesh, Aditya and Clark, Aidan and Ostrow, AJ and Welihinda, Akila and Hayes, Alan and Radford, Alec and others},
  journal={arXiv preprint arXiv:2410.21276},
  year={2024}
}

@misc{openai2023gpt4,
  title={GPT-4v System Card},
  author={OpenAI},
  year={2023},
  howpublished={\url{https://openai.com/index/gpt-4v-system-card/}},
}

@misc{openai2023dalle3,
  title={DALL·E 3 System Card},
  author={OpenAI},
  year={2023},
  howpublished={\url{https://openai.com/research/dall-e-3-system-card}},
}

@inproceedings{kumari2023multi,
  title={Multi-concept customization of text-to-image diffusion},
  author={Kumari, Nupur and Zhang, Bingliang and Zhang, Richard and Shechtman, Eli and Zhu, Jun-Yan},
  booktitle={Proceedings of the IEEE/CVF conference on computer vision and pattern recognition},
  pages={1931--1941},
  year={2023}
}

@online{Midjourney,
  author = {Midjourney},
  title = {Midjourney},
  url = {https://www.midjourney.com/},
  note = {Accessed: [Current Date, e.g., 2025-10-13]},
  year = {2022} ,
}

@article{wang2024ms,
  title={Ms-diffusion: Multi-subject zero-shot image personalization with layout guidance},
  author={Wang, Xierui and Fu, Siming and Huang, Qihan and He, Wanggui and Jiang, Hao},
  journal={arXiv preprint arXiv:2406.07209},
  year={2024}
}

@article{kang2025flux,
  title={Flux Already Knows--Activating Subject-Driven Image Generation without Training},
  author={Kang, Hao and Fotiadis, Stathi and Jiang, Liming and Yan, Qing and Jia, Yumin and Liu, Zichuan and Chong, Min Jin and Lu, Xin},
  journal={arXiv preprint arXiv:2504.11478},
  year={2025}
}
}


\end{document}